\documentclass[pdflatex,sn-mathphys,Numbered]{sn-jnl}

\usepackage{graphicx}%
\usepackage{multirow}%
\usepackage{amsmath,amssymb,amsfonts}%
\usepackage{amsthm}%
\usepackage{mathrsfs}%
\usepackage[title]{appendix}%
\usepackage{xcolor}%
\usepackage{colortbl}% provides \cellcolor for highlighted table entries
\usepackage{textcomp}%
\usepackage{manyfoot}%
\usepackage{booktabs}%
\usepackage{algorithm}%
\usepackage{algorithmicx}%
\usepackage{algpseudocode}%

\definecolor{RedOrange}{RGB}{255,180,120}
\definecolor{Peach}{RGB}{255,225,200}

\begin{document}

\title[RxGS: Relightable Gaussian Stylization via BRDF Decomposition]{RxGS: Relightable Gaussian Stylization via BRDF Decomposition}

\author[1]{\fnm{Wanrong} \sur{Hong}}
\author[2]{\fnm{Liu} \sur{Jiang}}
\author[3]{\fnm{Mukesh} \sur{Prasad}}
\author*[1]{\fnm{Jun} \sur{Li}}\email{jun.li@uts.edu.au}

\affil[1]{\orgdiv{School of Computer Science}, \orgname{University of Technology Sydney}, \city{Sydney}, \country{Australia}}
\affil[2]{\orgdiv{Department of Civil and Environmental Engineering}, \orgname{Stanford University}, \city{Stanford}, \country{United State}}
\affil[3]{\orgdiv{Australian Artificial Intelligence Institute}, \orgname{University of Technology Sydney}, \city{Sydney}, \postcode{2007}, \country{Australia}}

\abstract{Reference-based 3D scene stylization transfers the appearance of a reference image onto a reconstructed 3D scene while preserving scene geometry. Recent methods perform this transfer by optimizing the appearance representation of neural scene models including Neural Radiance Fields (NeRF) and 3D Gaussian Splatting (3DGS). These methods optimize an RGB appearance representation, in which a single per-point colour combines the effects of material reflectance, environment illumination, together with the shading it produces, and the viewing direction. When stylization modifies this combined colour, the illumination effects present during stylization are baked into the stored appearance, so the stylized scene cannot be rendered under different illumination. To address this limitation, we present RxGS, a relightable 3D scene stylization framework built on a Bidirectional Reflectance Distribution Function (BRDF)-aware 3D Gaussian representation. Rather than modifying a per-point RGB colour, RxGS optimizes only the diffuse albedo field recovered by inverse rendering, while leaving roughness, metallicity, surface normals, and the estimated environment lighting unchanged. As a result, the stylized scene keeps all the components a physically-based renderer needs other than the edited albedo, and can therefore still be relit. RxGS optimizes the albedo field using a nearest-neighbour feature matching loss together with a spatial smoothness prior over Gaussian centres. We evaluate the method using a Geometry Consistency Cost (GCC) based on optical-flow warping error. RxGS achieves the lowest GCC on all evaluated NeRF-Synthetic scenes and remains competitive on real-captured scenes, while additionally supporting relightable stylization, which existing reference-based 3D scene stylization methods do not provide.}


\maketitle

\section{Introduction}\label{sec:intro}

Reference-based 3D scene stylization transfers the appearance of an artistic reference image onto a reconstructed 3D scene while keeping the rendered result geometrically consistent across viewpoints. Recent neural rendering methods, in particular Neural Radiance Fields (NeRF) \cite{mildenhall2021nerf} and 3D Gaussian Splatting (3DGS) \cite{kerbl20233dgs}, produce high-quality novel views and have prompted a growing body of 3D stylization work. These methods stylize a scene by modifying the RGB appearance representation that the neural scene model has learned. In many applications a scene must be stylized once and then rendered under several different lighting conditions without repeating the stylization. This requires the stylized scene to keep a representation that can be relit after the appearance transfer, a requirement the RGB representation does not meet, as the next paragraph argues.

\begin{figure}[t]
    \centering
    \includegraphics[width=0.7\linewidth]{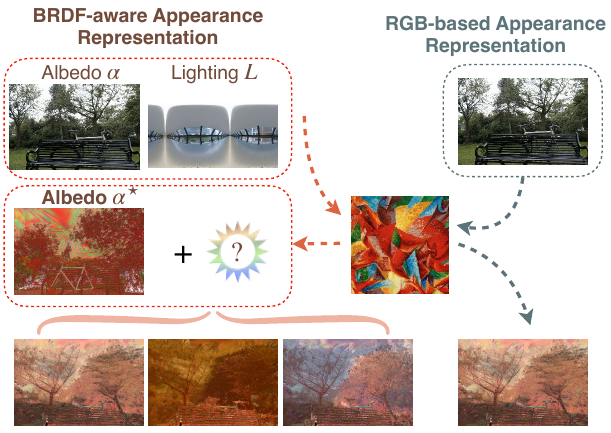}
    \caption{Comparison between stylization on an RGB appearance representation and stylization on a BRDF-aware scene representation. Existing methods transfer appearance onto a per-point RGB colour, which already combines material reflectance and illumination. The illumination present during stylization is therefore written into the stored colour, and the result no longer changes when the lighting changes. RxGS instead modifies only the diffuse albedo and leaves the estimated lighting representation unchanged, so the stylized albedo can be rendered under different illumination conditions, producing relightable stylized scenes without re-running the stylization.}
    \label{fig:relightVSRGB}
\end{figure}

An RGB appearance representation stores, at each point, one colour that already combines material reflectance, the environment illumination together with the shading it produces, and view-dependent effects. Because these factors are not stored separately, a stylization process that edits this colour necessarily edits the illumination contribution together with the intrinsic material. The illumination present during stylization is thereby written into the stored appearance: when the scene is later rendered under a different illumination, the stored colour does not change, so no new shading or shadows appear. As illustrated in Fig.~\ref{fig:relightVSRGB}, stylization applied to an RGB appearance representation fixes the artistic appearance and the illumination contribution together, which prevents the result from responding to novel lighting.

To remove this restriction we build on recent inverse rendering methods for 3D Gaussian representations. Such methods \cite{liang2024gsir} recover separate scene attributes, namely diffuse albedo, material properties, and environment illumination, and so support relightable rendering and material editing. These separated representations have not previously been used for reference-based 3D scene stylization.

We present RxGS, a relightable 3D scene stylization framework built on a Bidirectional Reflectance Distribution Function (BRDF)-aware Gaussian scene representation. Given a reconstructed scene, RxGS first applies inverse rendering decomposition to recover, per Gaussian, the diffuse albedo $\mathbf{a}_i$, the roughness $r_i$, the metallicity $m_i$, the surface normal $\mathbf{n}_i$, and one global environment lighting $L$. Stylization then optimizes only the field of diffuse albedos $\{\mathbf{a}_i\}$ and leaves $\{r_i, m_i, \mathbf{n}_i\}$ and $L$ and all geometry unchanged. Relightability follows from this bookkeeping rather than from any claim that the albedo alone fixes the rendered colour. A physically-based renderer consumes the geometry, the per-Gaussian material attributes $\{\mathbf{a}_i, r_i, m_i, \mathbf{n}_i\}$, and the environment lighting $L$. RxGS changes only the albedo coefficients $\{\mathbf{a}_i^{\star}\}$ and holds every other input fixed, so all non-albedo inputs the renderer needs remain intact, and the same renderer can be re-run under any chosen $L$ to produce a relit image of the stylized scene. To optimize the albedo field, RxGS uses a nearest-neighbour feature matching (NNFM) loss in the VGG feature space together with a spatial smoothness term over Gaussian centres.

We evaluate RxGS on synthetic and real-captured scenes, measuring cross-view geometric consistency with the Geometry Consistency Cost (GCC) defined in Section~\ref{sec:experiments}. RxGS attains the lowest GCC on every evaluated NeRF-Synthetic scene and remains competitive on real-captured scenes, while keeping a relightable scene representation after stylization. Our contributions are the following.

\begin{itemize}
    \item We present RxGS, a relightable 3D scene stylization framework based on a BRDF-aware Gaussian scene representation.
    \item We formulate stylization as optimization of the diffuse albedo field alone in a logit parameterization, regularized by a smoothness term on a $k$-nearest-neighbour graph over Gaussian centres, so that artistic appearance transfer leaves the material and illumination parameters required for relighting unchanged.
    \item We evaluate on synthetic and real-captured datasets, reporting cross-view geometric consistency (GCC) that is competitive with prior stylization methods together with relightable stylization that prior methods do not provide.
\end{itemize}

\section{Related work}\label{sec:related}

\subsection{3D Scene Representation and Rendering}

A 3D scene representation encodes scene geometry and appearance in a compact, learnable form that supports downstream tasks such as novel view synthesis, scene editing, relighting, and scene understanding. NeRF \cite{mildenhall2021nerf} introduced an implicit neural representation that parameterizes volumetric density and radiance with multilayer perceptrons, achieving novel view synthesis from multi-view observations. NeRF requires dense volumetric sampling during both training and rendering, which is computationally expensive. To reduce this cost, later work introduced voxel grids, tensor factorizations, and multi-resolution encodings, including DVGO \cite{Sun22dvgo}, Plenoxels \cite{Fridovich2022plenoxels}, TensoRF \cite{Chen22TensoRF}, and Instant-NGP \cite{muller2022instant}. As scene editing and stylization tasks developed, several methods adopted surface-based representations that give a more explicit geometric correspondence than a purely volumetric field, including VolSDF \cite{Yariv2021VolSDF}, NeuS \cite{Wang2021neus}, NeuMesh \cite{Yang2022NeuMesh}, and NeRF2Mesh \cite{Tang2023NeRF2Mesh}.

Among explicit representations, 3D Gaussian Splatting (3DGS) \cite{kerbl20233dgs} represents a scene with a set of anisotropic Gaussian primitives. Unlike volumetric methods that rely on dense ray sampling, 3DGS renders through differentiable Gaussian rasterization, which is fast enough for real-time rendering while still reconstructing scene geometry and appearance. Following 3DGS, many extensions adapt the representation to downstream tasks, including scene editing, relighting, and inverse rendering. The inverse rendering line, which recovers separate material and lighting attributes for each Gaussian, is the one RxGS builds upon, and we review it next.

\subsection{Inverse Rendering with 3D Gaussian Representations}

Inverse rendering recovers physically meaningful scene properties, namely geometry, material attributes, and illumination, from image observations. Several methods built this decomposition into neural rendering frameworks. NeRFactor \cite{zhang2021Nerfactor} factorizes scene appearance into geometry, reflectance, visibility, and illumination for relightable rendering. PhySG \cite{zhang2021physg} uses spherical Gaussian representations of the BRDF and of the illumination to put physically-based shading into neural rendering. TensorIR \cite{Jin2023tensorir} introduces a tensor-based inverse rendering framework that makes scene decomposition and relightable rendering more efficient.

After 3DGS, inverse rendering was extended to Gaussian representations by attaching material, normal, and illumination attributes to the Gaussian primitives. GaussianShader \cite{jiang2024gaussianshader} adds a shading function to Gaussian rendering and estimates geometry-aware normals to improve reflective surfaces. Relightable 3D Gaussian \cite{gao2024relightable} attaches BRDF attributes and incident lighting to each Gaussian and renders through differentiable physically-based rendering. GS-IR \cite{liang2024gsir} formulates inverse rendering on a BRDF-aware Gaussian representation by jointly estimating scene geometry, material properties, and environment illumination from multi-view observations; RxGS takes the decomposition produced by GS-IR as its input.

\subsection{3D Scene Stylization}

Neural style transfer was first developed for images \cite{gatys2016image, huang2017adain, Doyle2019LaplacianPyramid, Xu2024GlobalLocalStyle}, where the goal is to synthesize an image that keeps the scene content of one input while matching the appearance of a reference style image. As 3D scene representations matured, stylization was extended from 2D images to 3D scenes. Before neural scene representations, 3D stylization operated on explicit representations such as meshes and point clouds \cite{hollein2022stylemesh, cao2020psnet, liu2018paparazzi}, manipulating surface textures, geometric attributes, or rendered appearance directly.

Neural scene representations let stylization act on a learned appearance representation, so that style is propagated consistently across viewpoints. Early NeRF-based methods modify the appearance field of a reconstructed scene while keeping the geometry fixed \cite{chiang2022stylizing, zhang2022arf, chen2024upst, nguyen2022snerf, goel2022styletrf, huang2022stylizednerf}. ARF \cite{zhang2022arf} optimizes the appearance field of a NeRF with a neural style transfer objective, giving multi-view consistent stylization, 
and ARF-plus \cite{li2025arfplus} improves stylization quality and training stability through stronger appearance representations and optimization strategies. Ref-NPR \cite{zhang2023refnpr} improves controllability by transferring appearance from reference images through geometry-aware feature correspondence, which enables localized stylization. LAE-NeRF \cite{radl2023laenerf} lets a user mark target regions in a neural scene representation and propagates the edit through neighbouring volumetric regions. SNeRF \cite{nguyen2022snerf} alternates between image-space stylization and scene reconstruction, transferring artistic appearance into the neural scene representation across iterations.

Because 3DGS associates appearance attributes directly with individual primitives, it makes appearance manipulation explicit. StylizedGS \cite{zhang2025stylizedgs} supports region-specific style transfer through semantic masks. ReGS \cite{mei2024ReGS} adapts 3DGS to reference-based stylization with a texture-guided mechanism that adjusts local Gaussian arrangements to reproduce texture detail from the reference. StyleSplat segments individual objects in a Gaussian scene and fine-tunes the selected Gaussians for object-level stylization. Further methods match segmented regions across views before transfer to keep the style consistent over the whole scene \cite{Liu25abcgs}, geometry-aware texture transfer \cite{liu2026gt2gs}, diffusion-based stylization and distillation \cite{yang2026fantacy3d}, and feed-forward or optimization-free stylization of Gaussian splats \cite{Liu2026Stylos, sablon2026optifree}. All of these methods, whether NeRF-based or Gaussian-based, edit a per-point RGB appearance, and so produce a stylized scene that stores one fixed colour per point and cannot be relit.

A smaller line of work stylizes a decomposed appearance representation rather than RGB colour. TANGO \cite{chen2022tango} performs text-driven stylization by editing BRDF attributes on a relightable mesh, which allows appearance editing under varying illumination. RxGS differs from this mesh-based, text-driven setting: it performs reference-image stylization directly on a BRDF-aware Gaussian representation, editing only the diffuse albedo field of the Gaussians and keeping the rest of the decomposition fixed.

\begin{figure*}[t!]
    \centering
    \includegraphics[width=\textwidth]{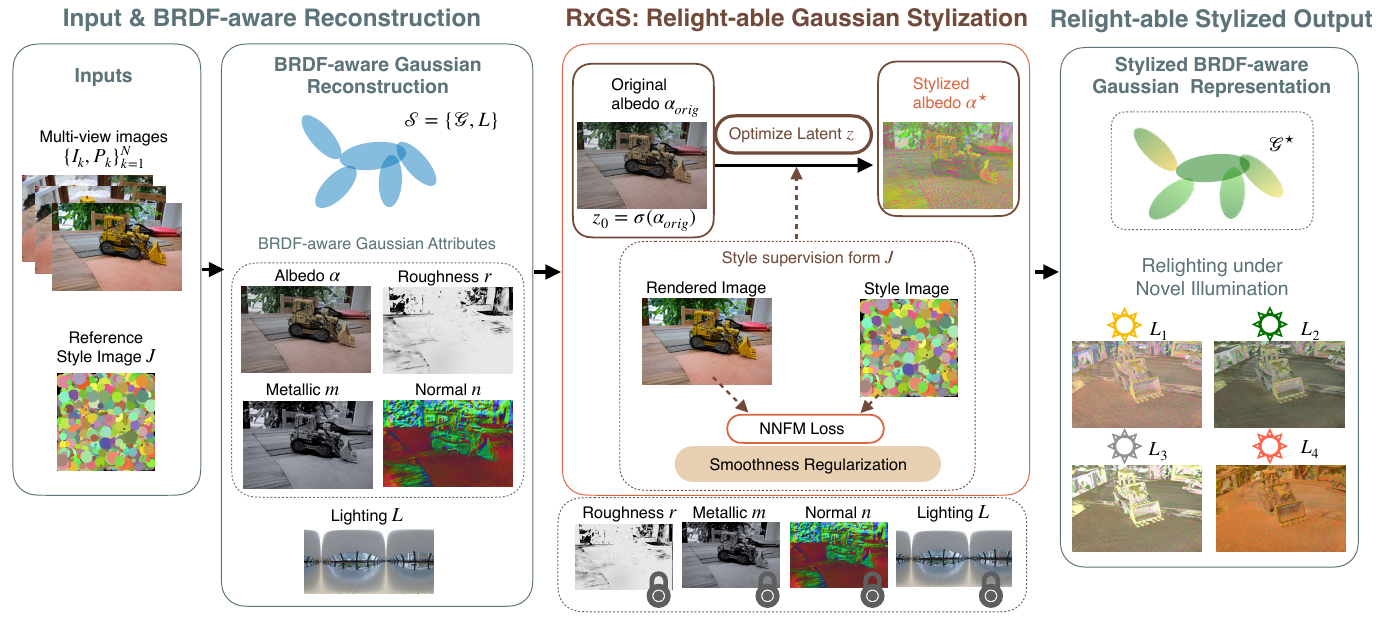}
    \caption{Overview of RxGS. Given posed multi-view images and a reference style image, a BRDF-aware Gaussian scene representation is first reconstructed with GS-IR, which separates the scene into geometry, diffuse albedo, material attributes (roughness and metallicity), surface normals, and one global environment lighting. RxGS stylizes by optimizing only the diffuse albedo field, with a nearest-neighbour feature matching (NNFM) loss and a spatial smoothness term, while leaving roughness, metallicity, surface normals, and the lighting unchanged. The stylized BRDF-aware representation therefore keeps its relightable property and can be rendered under novel illumination without further optimization.}
    \label{fig:main-workflow}
\end{figure*}

\section{Method}\label{sec:method}

The task is to modify the appearance of a 3D scene, reconstructed from posed multi-view images $\{I_k, P_k\}_{k=1}^K$, so that it matches the style of a reference image $J$, while keeping both the scene geometry and the physical rendering components that are needed to render the scene under novel illumination.

\subsection{Gaussian Representation}

\subsubsection{3D Gaussian Splatting}

3D Gaussian Splatting (3DGS) \cite{kerbl20233dgs} represents a scene with a set of anisotropic 3D Gaussian primitives placed in 3D space. Each Gaussian is parameterized by a centre $\mathbf{x}_i \in \mathbb{R}^3$, a scaling vector $\mathbf{s}_i \in \mathbb{R}^3$, a rotation quaternion $\mathbf{q}_i \in \mathbb{R}^4$, an opacity $\alpha_i \in [0,1]$, and appearance-related attributes. The scaling vector and rotation quaternion together define the Gaussian covariance. The trainable parameters of the $i$-th Gaussian are
\begin{align}
    G_i = \{\mathbf{x}_i, \mathbf{s}_i, \mathbf{q}_i, \alpha_i, \phi_i\},
\end{align}
where $\phi_i$ denotes the appearance representation of the Gaussian.

Because the Gaussian primitives are defined in 3D space, the appearance attributes of a Gaussian are shared by every viewpoint that observes the same spatial region. This gives a fixed correspondence between scene geometry and appearance during rendering, so a change to a Gaussian's appearance attributes affects all views of that region in the same way, which is useful for keeping a stylization consistent across viewpoints.

\subsubsection{BRDF-aware Gaussian Representation}

GS-IR \cite{liang2024gsir} refines the 3DGS appearance attribute $\phi_i$ into a BRDF-aware decomposition consisting of diffuse albedo, roughness, metallicity, and surface normal, while jointly estimating a global environment lighting $L$, represented as an environment map, that is, a function on the unit sphere of incident directions giving the incoming radiance from each direction. Specifically,
\begin{align}
    \phi_i = \{\mathbf{a}_i, r_i, m_i, \mathbf{n}_i\},
    \label{eq:MaterialRepre}
\end{align}
where $\mathbf{a}_i \in [0,1]^3$ is the diffuse albedo of the $i$-th Gaussian, $r_i \in [0,1]$ is the roughness, $m_i \in [0,1]$ is the metallicity, and $\mathbf{n}_i \in \mathbb{R}^3$ is the surface normal.

In an RGB appearance representation a single per-point colour already combines material reflectance, the illumination together with the shading it produces, and view-dependent effects. The decomposition above instead stores the albedo $\mathbf{a}_i$ as a separate per-Gaussian coefficient, alongside $r_i$, $m_i$, $\mathbf{n}_i$, and the global lighting $L$. 
Although the diffuse albedo is represented separately, it still influences the colour of specular reflections under the metallic-roughness BRDF when metallicity is non-zero. Therefore, modifying the albedo changes both the diffuse appearance and the tint of specular highlights. However, the remaining BRDF attributes, including roughness, metallicity, surface normals, and environment lighting, remain unchanged throughout stylization. Consequently, the scene retains the complete set of rendering parameters required for relightable rendering.

\subsection{RxGS: Relightable Gaussian Stylization}

RxGS takes as input posed multi-view images $\{I_k, P_k\}_{k=1}^K$ and a reference style image $J$. Let the BRDF-aware scene representation be
\[
\mathcal{S} = \{\mathcal{G}, L\},
\]
where $\mathcal{G} = \{G_i\}_{i=1}^{N}$ is the set of $N$ Gaussian primitives and $L$ is the estimated environment lighting. RxGS optimizes the diffuse albedo field while keeping all remaining BRDF attributes, geometry, and environment lighting fixed.

We separate the parameters of each Gaussian into geometry-related and appearance-related parts,
\begin{align}
    G_i = \{G_i^{\mathrm{geo}}, \phi_i\},
\end{align}
where $G_i^{\mathrm{geo}} = \{\mathbf{x}_i, \mathbf{s}_i, \mathbf{q}_i, \alpha_i\}$. The stylized appearance decomposition is
\begin{align}
    \phi_i^{\star} = \{\mathbf{a}_i^{\star}, r_i, m_i, \mathbf{n}_i\},
    \label{eq:OptimizedMaterial}
\end{align}
where only the diffuse albedo $\mathbf{a}_i^{\star}$ is optimized; the geometry attributes $G_i^{\mathrm{geo}}$, the roughness $r_i$, the metallicity $m_i$, the surface normal $\mathbf{n}_i$, and the environment lighting $L$ are all held fixed. Fig.~\ref{fig:main-workflow} gives an overview.

Existing reference-based 3D scene stylization methods perform style transfer on rendered RGB images. Given a rendered image $I$, the stylized result is
\begin{align}
    \tilde{I} = h^J(I),
\end{align}
where $h^J(\cdot)$ is a style transfer operator conditioned on the reference style image $J$. Because the rendered RGB image already combines material reflectance, shading, illumination, and view-dependent effects at each pixel, $h^J$ modifies all of these together. The illumination present in $I$ is thereby encoded in $\tilde{I}$. When the scene is later rendered under a different illumination, the stored stylized colour does not change, so the result shows no new illumination-dependent variation.

RxGS instead modifies only the diffuse albedo field $\{\mathbf{a}_i^{\star}\}$ and leaves $\{r_i, m_i, \mathbf{n}_i\}$ and $L$ unchanged. Since only the albedo coefficient is changed and $r_i$, $m_i$, $\mathbf{n}_i$, $L$, and the geometry are held fixed, every rendering input required by the physically-based renderer other than the albedo is intact. The same renderer can therefore be run again under any chosen $L$, and it produces illumination-dependent variation when relit.

\subsubsection{Albedo Parameterization}

Because only the diffuse albedo is optimized, the optimization must keep its value in the valid reflectance range $[0,1]^3$. Optimizing $\mathbf{a}_i^{\star}$ directly could leave this range. RxGS therefore parameterizes the stylized albedo $\mathbf{a}_i^{\star} \in [0,1]^3$ through an unconstrained latent variable $\mathbf{z}_i \in \mathbb{R}^3$ in logit space. Let $\mathbf{a}_{i,\mathrm{orig}}$ denote the original albedo of Gaussian $G_i$ from the BRDF-aware decomposition. The component-wise inverse sigmoid $\sigma^{-1}$ diverges at $0$ and $1$, so before applying it we clamp each channel of $\mathbf{a}_{i,\mathrm{orig}}$ to $[\varepsilon, 1-\varepsilon]$ with $\varepsilon = 10^{-6}$. We initialize
\begin{align}
    \mathbf{z}_{i,0} = \sigma^{-1}\!\big(\mathrm{clamp}(\mathbf{a}_{i,\mathrm{orig}})\big),
\end{align}
and recover the stylized albedo as
\begin{align}
    \mathbf{a}_i^{\star} = \sigma(\mathbf{z}_i),
\end{align}
where $\sigma$ is the component-wise sigmoid function and $\sigma^{-1}$ its inverse. With this parameterization $\mathbf{a}_i^{\star}$ stays in $[0,1]^3$ for every value of $\mathbf{z}_i$, and at initialization $\mathbf{a}_i^{\star}$ equals the clamped recovered albedo.

\subsubsection{Rendering Strategy}

The style supervision is computed on images rendered from the stylized scene. Let
\[
\mathcal{G}^{\star} = \{G_i^{\star}\}_{i=1}^{N}, \qquad G_i^{\star} = \{G_i^{\mathrm{geo}}, \phi_i^{\star}\},
\]
denote the stylized Gaussian scene. The direct strategy renders the stylized albedo field without illumination,
\begin{align}
    \hat{I}_k = R_a(\mathcal{G}^{\star}, P_k),
\end{align}
where $R_a(\cdot)$ is albedo rendering, which composites the per-Gaussian albedo $\mathbf{a}_i^{\star}$ along each ray without applying any shading. Style supervision is then applied directly in the albedo domain.

We additionally consider a randomized-illumination strategy. Here, at optimization step $t$, a lighting $L^{(t)}$ is randomly sampled and the stylized scene is rendered under it with a Lambertian, that is diffuse-only, shading model,
\begin{align}
    \hat{I}_k = R(\mathcal{G}^{\star}, P_k, L^{(t)}),
\end{align}
where $R(\cdot)$ is a renderer that applies the sampled lighting. Section~\ref{sec:ablation} compares the two strategies.

\subsubsection{Style Loss}

Following ARF \cite{zhang2022arf}, we use the nearest-neighbour feature matching (NNFM) loss as the style objective. Given a rendered image $\hat{I}_k$ and the reference style image $J$, the style loss is $\mathcal{L}_{\mathrm{nnfm}}(\hat{I}_k, J)$. It extracts feature maps from a pre-trained VGG-16 network for both images, and for each feature vector of $\hat{I}_k$ it finds the most similar feature vector of $J$ by cosine similarity, then penalizes the cosine distance to that match. Following ARF, the features are taken from the \textit{conv3} block of the VGG-16 network.

To control the trade-off between content preservation and stylization strength, RxGS optionally incorporates a content loss. The stylization objective is
\begin{align}
    \mathcal{L}_{\mathrm{style}} = \mathcal{L}_{\mathrm{nnfm}} + \lambda_c \mathcal{L}_{\mathrm{content}},
\end{align}
where $\lambda_c$ weights the content term. Unless stated otherwise we set $\lambda_c = 0$, so stylization is driven by the NNFM objective alone.

\subsubsection{Spatial Smoothness Prior on the Gaussian Neighbourhood Graph}
\label{sec:method-knn}

To make the stylized albedo vary smoothly over the scene, we add a smoothness term on the optimized albedo field. The albedo is stored on the Gaussian primitives rather than on a regular image grid, so the smoothness prior is defined on a $k$-nearest-neighbour graph built from the Gaussian centres. For a Gaussian $G_i$, its neighbourhood $\mathcal{N}_k(i)$ is the set of the $k$ Gaussians whose centres have the smallest Euclidean distance to $\mathbf{x}_i$. The spatial gradient of the stylized albedo field at $G_i$ is the average, over the neighbourhood, of the albedo difference weighted by the inverse Euclidean distance between Gaussian centres,
\begin{align}
    \nabla_{\mathrm{spatial}} \mathbf{a}_i^{\star}
    = \frac{1}{\lvert \mathcal{N}_k(i) \rvert}
      \sum_{j \in \mathcal{N}_k(i)}
      \frac{\mathbf{a}_i^{\star} - \mathbf{a}_j^{\star}}
           {\lVert \mathbf{x}_i - \mathbf{x}_j \rVert_2 + \varepsilon},
    \label{eq:spatial-grad}
\end{align}
where $\varepsilon = 10^{-6}$ avoids division by zero and $k = 16$ in all experiments. The smoothness loss sums the $L_1$ norm of this gradient over all Gaussians,
\begin{align}
    \mathcal{L}_{\mathrm{smooth}}
    = \sum_i \lVert \nabla_{\mathrm{spatial}} \mathbf{a}_i^{\star} \rVert_1 .
\end{align}
The final objective combines the style and smoothness terms,
\begin{align}
    \mathcal{L} = \mathcal{L}_{\mathrm{style}} + \lambda_s \mathcal{L}_{\mathrm{smooth}},
\end{align}
where $\lambda_s = 10^{-3}$.

\section{Experiments}\label{sec:experiments}

We evaluate RxGS from two angles: cross-view geometric consistency and relightable stylization. We first compare RxGS with existing reference-based 3D scene stylization methods using a geometric consistency metric, which measures how well scene structure is preserved across viewpoints after stylization. Because the existing methods do not support relighting, we then evaluate the relightable stylization of RxGS by rendering stylized scenes under different illumination conditions and examining how the appearance responds.

\subsection{Experimental Setup}

\paragraph{Datasets.}
We use NeRF-Synthetic \cite{mildenhall2021nerf} and NeRF-LLFF \cite{mildenhall2019local} for the geometric consistency comparison, and add TensorIR \cite{Jin2023tensorir} and Mip-NeRF 360 \cite{Barron22_Mip360} for the relightable stylization experiments. NeRF-Synthetic \cite{mildenhall2021nerf} is a synthetic multi-view dataset of eight object-centric scenes, each with 100 training images and 200 test images rendered at $800 \times 800$, with camera poses and foreground masks. NeRF-LLFF \cite{mildenhall2019local} is a real-world forward-facing multi-view dataset captured by handheld cameras with calibrated poses; it contains more complex geometry, occlusions, and appearance variation than NeRF-Synthetic, and we downsample the input images by a factor of four, following prior work \cite{zhang2023refnpr, Fridovich2022plenoxels, Tang2023NeRF2Mesh}. TensorIR \cite{Jin2023tensorir} is a multi-view dataset for inverse rendering and relighting, with diverse materials and illumination and with ground-truth geometry, material, and lighting; we use its \textit{Lego} and \textit{Hotdog} scenes for relightable stylization. Mip-NeRF 360 \cite{Barron22_Mip360} is a real-world multi-view dataset of unbounded indoor and outdoor scenes captured over 360-degree viewpoints, with larger extents and more complex structure than NeRF-Synthetic and LLFF; we use its \textit{Garden}, \textit{Kitchen}, and \textit{Bicycle} scenes.

\paragraph{Implementation Details.}
We take the BRDF-aware Gaussian representation reconstructed by GS-IR \cite{liang2024gsir} as the input to RxGS. Stylization optimizes the diffuse albedo while keeping the estimated geometry, roughness, metallicity, surface normals, and environment lighting fixed. Following ARF \cite{zhang2022arf}, we use the NNFM loss with a pre-trained VGG-16 network and take features from the \textit{conv3} block. Each scene is optimized for 1000 iterations.

\paragraph{Comparison Methods.}
We compare RxGS with ARF \cite{zhang2022arf}, Ref-NPR \cite{zhang2023refnpr}, and StylizedGS \cite{zhang2025stylizedgs}. Ref-NPR requires a reference image rather than a style image, so we generate its reference images with the AdaIN style transfer model \cite{huang2017adain} from the same style images the other methods use. StylizedGS is included only on the LLFF scenes, because the original paper does not report NeRF-Synthetic results.

\paragraph{Evaluation Protocol.}
To reduce bias toward a particular artistic appearance, we evaluate each scene with three different reference styles, and for the geometric consistency comparison we average the reported results over the three styles. For relighting, NeRF-Synthetic and LLFF scenes are rendered under several illumination conditions by rotating the estimated environment map while keeping all stylization parameters fixed. We additionally evaluate relightable stylization on TensorIR and Mip-NeRF 360 scenes to examine the behaviour of stylized scenes under varying illumination across different scene types.

\paragraph{Metric.}
To measure cross-view geometric consistency we use the Geometry Consistency Cost (GCC), a per-pixel optical-flow warping error used in earlier video and 3D stylization work \cite{ruder2016, zhang2023refnpr}. Given two stylized renderings $\hat{I}_k$ and $\hat{I}_{k-1}$ from nearby viewpoints, we estimate the optical flow from $\hat{I}_{k-1}$ to $\hat{I}_k$ with the pre-trained RAFT model \cite{teed2020raft}, warp $\hat{I}_{k-1}$ to the viewpoint of $\hat{I}_k$ using this flow, and compute the mean per-pixel error between $\hat{I}_k$ and the warped frame over the pixels that the warp marks as visible. Averaging this error over consecutive frame pairs of a scene gives its GCC; a lower GCC indicates higher cross-view geometric consistency.

\begin{table}[t]
\centering
\caption{Geometry Consistency Cost (GCC) for stylization of the NeRF-Synthetic dataset \cite{mildenhall2021nerf}. Lower is better. The best result in each column is highlighted in orange. StylizedGS \cite{zhang2025stylizedgs} does not report results on NeRF-Synthetic and is therefore omitted from this comparison.}
\label{tab:synthetic_result}
\begin{tabular}{l|c|c|c|c|c|c|c|c}
\hline
Method & Lego & Ship & Chair & Mic & Drums & Ficus & Hotdog & Materials \\\hline
ARF \cite{zhang2022arf} & 0.0038 & 0.017 & 0.0061 & 0.0024 & 0.0028 & 0.0024 & 0.0043 & 0.0036 \\
Ref-NPR \cite{zhang2023refnpr} & 0.0045 & 0.0085 & 0.0034 & 0.0015 & 0.0035 & 0.0015 & 0.0047 & 0.0042 \\
RxGS & \cellcolor{RedOrange} 0.0017 & \cellcolor{RedOrange} 0.0045 & \cellcolor{RedOrange} 0.0028 & \cellcolor{RedOrange} 0.0004 & \cellcolor{RedOrange} 0.0017 & \cellcolor{RedOrange} 0.0005 & \cellcolor{RedOrange} 0.0029 & \cellcolor{RedOrange} 0.0011 \\
\hline
\end{tabular}
\end{table}

\begin{table}[t]
\centering
\caption{Geometry Consistency Cost (GCC) for stylization of the LLFF dataset \cite{mildenhall2019local}. Lower is better. The best and second-best results are highlighted in orange and peach, respectively.}
\label{tab:llff_result}
\begin{tabular}{l|c|c|c|c|c|c|c|c}
\hline
Method & Fern & Fortress & Horns & Leaves & Orchids & Room & Trex & Flower \\\hline
ARF \cite{zhang2022arf} & 0.0084 & 0.0168 & 0.0129 & 0.0111 & 0.0083 & 0.0093 & 0.0117 & 0.0126 \\
Ref-NPR \cite{zhang2023refnpr} & \cellcolor{Peach} 0.0064 & 0.0111 & 0.0086 & \cellcolor{Peach} 0.0071 & \cellcolor{Peach} 0.0077 & 0.0071 & 0.0085 & \cellcolor{Peach} 0.0082 \\
StylizedGS \cite{zhang2025stylizedgs} & \cellcolor{RedOrange} 0.0056 & \cellcolor{RedOrange} 0.0054 & \cellcolor{RedOrange} 0.0074 & \cellcolor{RedOrange} 0.0052 & \cellcolor{RedOrange} 0.0051 & \cellcolor{RedOrange} 0.0068 & \cellcolor{RedOrange} 0.0064 & \cellcolor{RedOrange} 0.0067 \\
RxGS & 0.0078 & \cellcolor{Peach} 0.0105 & \cellcolor{Peach} 0.0085 & 0.0084 & 0.0084 & 0.0102 & \cellcolor{Peach} 0.0080 & 0.0113 \\
\hline
\end{tabular}
\end{table}

\begin{figure*}[t]
    \centering
    \includegraphics[width=0.9\textwidth]{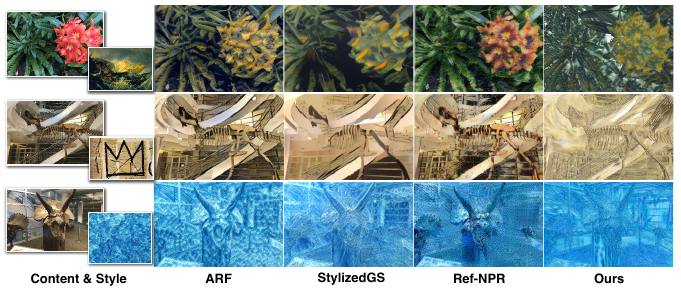}
    \caption{Qualitative comparison of stylization on LLFF scenes. Each row shows a content scene and its style reference image (left), then the results of ARF, StylizedGS, Ref-NPR, and RxGS. For RxGS the stylized albedo field is shown by flat-albedo rendering. All methods receive the same content and style inputs.}
    \label{fig:comp_llff}
\end{figure*}

\begin{figure}[t]
    \centering
    \includegraphics[width=0.7\linewidth]{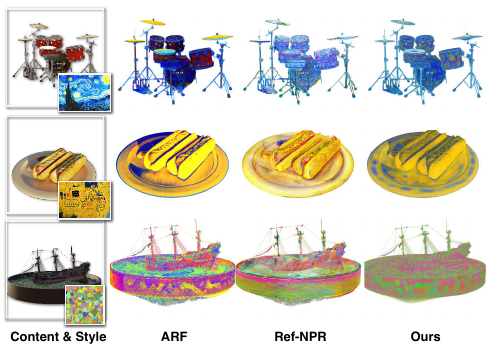}
    \caption{Qualitative comparison of stylization on NeRF-Synthetic scenes. Each row shows a content scene and its style reference image (left), then the results of ARF, Ref-NPR, and RxGS. For RxGS the stylized albedo field is shown by flat-albedo rendering.}
    \label{fig:comp_syn}
\end{figure}

\subsection{Cross-view Geometry Consistency}

\subsubsection{Quantitative Comparison}

Using the GCC metric above, we evaluate the cross-view geometric consistency of the scenes stylized by RxGS and compare with ARF, Ref-NPR, and StylizedGS. Table~\ref{tab:synthetic_result} and Table~\ref{tab:llff_result} report the GCC values on NeRF-Synthetic and LLFF.

On NeRF-Synthetic, RxGS attains the lowest GCC in every scene. The margin over ARF and Ref-NPR is largest on \textit{Lego}, \textit{Mic}, \textit{Ficus}, and \textit{Ship}. On LLFF, RxGS has a lower GCC than ARF and is comparable with Ref-NPR and StylizedGS on most scenes. Despite the greater complexity of the real-captured scenes, RxGS stays on par with these methods while also keeping a relightable scene representation, which they do not.

\begin{figure*}[t]
    \centering
    \includegraphics[width=0.9\textwidth]{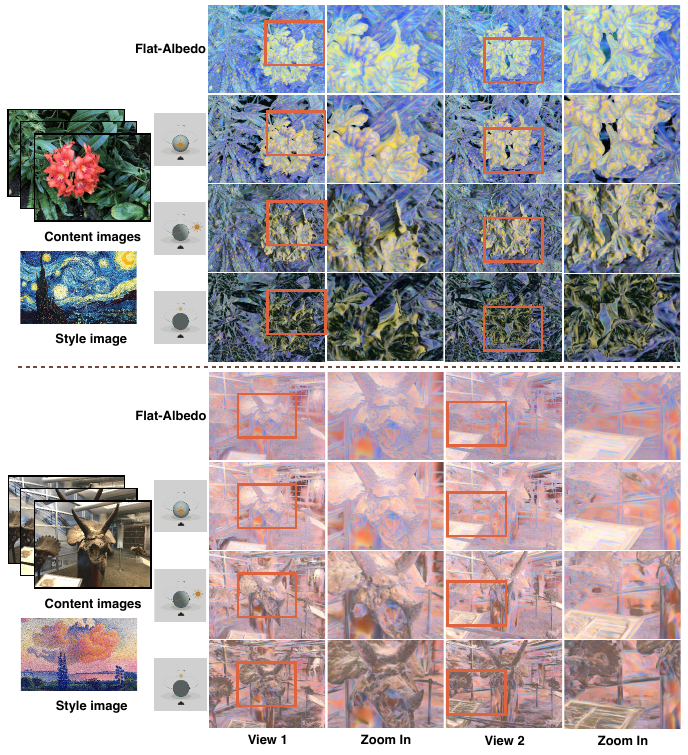}
    \caption{Relightable stylization on LLFF scenes. The same stylized representation is rendered under different illumination. The first row is albedo rendering without illumination; the remaining rows are renderings under the original environment lighting and under the environment map rotated by 90 and 180 degrees. Changing the illumination produces different shading and shadows while the stylized appearance stays consistent across viewpoints.}
    \label{fig:relighting_llff}
\end{figure*}

\subsubsection{Qualitative Comparison}

Fig.~\ref{fig:comp_llff} and Fig.~\ref{fig:comp_syn} show qualitative stylization results on NeRF-Synthetic and LLFF. On both datasets, RxGS transfers the colour distribution and texture pattern of the reference style image while keeping the scene structure; these results hold on synthetic scenes (Fig.~\ref{fig:comp_syn}) and on real-captured scenes (Fig.~\ref{fig:comp_llff}) alike. The RxGS results are rendered from the optimized albedo field without illumination, which lets the transferred appearance be examined independently of scene lighting. Relightable renderings under varying illumination follow in the next section.

\begin{figure*}[t]
    \centering
    \includegraphics[width=0.9\textwidth]{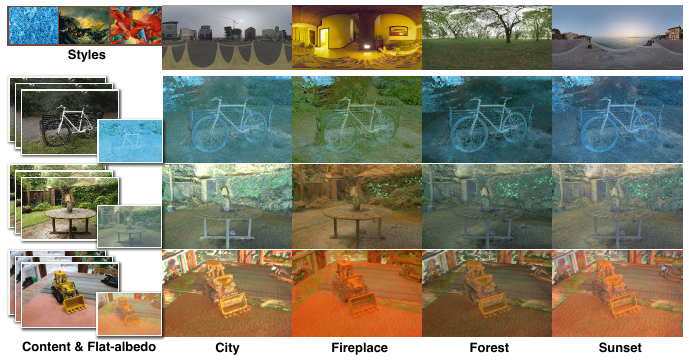}
    \caption{Relightable stylization on Mip-NeRF 360 under different environment lighting. For each scene the leftmost columns show the style image, the content image, and the stylized albedo rendering. The remaining columns render the same stylized scene under four environment maps (\textit{City}, \textit{Fireplace}, \textit{Forest}, and \textit{Sunset}). All results come from the same stylized representation without further optimization. The stylized scene responds to the illumination change, producing different lighting-dependent appearance while the stylized texture and the scene structure stay fixed.}
    \label{fig:relighting_360}
\end{figure*}

\subsection{Relightable Stylized Rendering}

Fig.~\ref{fig:relighting_llff} shows relighting results on stylized LLFF scenes under different illumination. As the illumination changes, the rendered appearance shows the corresponding shading while keeping the stylized texture and colour distribution. In the \textit{Flower} scene, rotating the dominant light source moves the shadows across the petals: under the 90-degree rotated lighting, shadows appear on the petals that face away from the incoming light. The \textit{Horns} scene behaves similarly, with shadow locations shifting as the lighting direction changes; when the illumination rotates toward the right of the scene, shadows appear on the left of the horns, and under back-facing illumination larger shadowed regions appear in the foreground and the overall appearance is darker. The stylized scene therefore remains responsive to illumination changes after stylization, producing different shading under different lighting while preserving the stylized texture and scene structure.

Fig.~\ref{fig:relighting_360} shows relighting results on stylized Mip-NeRF 360 scenes under four environment maps. The stylized appearance stays the same, but the four maps produce distinct illumination-dependent appearance. The City environment gives brighter renderings with stronger overall illumination, the Fireplace environment adds warmer colour tones, the Forest environment adds a green tint, and the Sunset environment gives a softer appearance with more moderate lighting transitions. These effects appear across the \textit{Bicycle}, \textit{Garden}, and \textit{Kitchen} scenes: the stylized texture and structural detail stay fixed while the rendered appearance follows the illumination of each environment map. RxGS therefore keeps a relightable scene representation after stylization, so the stylized scene responds to different lighting environments without re-running the stylization.

\subsection{Ablation Study}\label{sec:ablation}

\subsubsection{Effect of the Style Supervision Strategy}

We compare two supervision strategies. The first renders the stylized albedo without illumination, so style supervision acts in the albedo domain. The second renders the stylized albedo under randomly sampled Lambertian illumination, which adds illumination-dependent shading during optimization.

The comparison uses eight NeRF-Synthetic scenes and two reference styles, giving 16 scene-style pairs. For one optimization run we define its loss reduction as $(\mathcal{L}_{\mathrm{init}} - \mathcal{L}_{\mathrm{final}}) / \mathcal{L}_{\mathrm{init}}$, where $\mathcal{L}_{\mathrm{init}}$ and $\mathcal{L}_{\mathrm{final}}$ are the stylization loss $\mathcal{L}_{\mathrm{style}}$ at the first and the last iteration. Direct albedo supervision gives a lower final stylization loss than randomized-illumination supervision on all 16 pairs. Averaged over the 16 pairs, the loss reduction is $32.2\%$ for direct albedo supervision and $20.1\%$ for randomized-illumination supervision.

We attribute this gap to the extra appearance variation that randomized illumination introduces. Because the lighting direction changes during optimization, the rendered appearance contains illumination-dependent shading that is unrelated to the target style. The style supervision signal is then less stable and feature matching in the VGG space is harder. Direct albedo supervision removes the illumination contribution, so the optimization works only on the intrinsic appearance transfer. We therefore use direct albedo supervision in all experiments unless stated otherwise.

\section{Conclusion}\label{sec:conclusion}

We presented RxGS, a relightable 3D scene stylization framework built on a BRDF-aware Gaussian scene representation. RxGS optimizes only the diffuse albedo field and leaves the roughness, metallicity, surface normals, and environment lighting fixed, so artistic appearance transfer does not alter the material or illumination parameters and the stylized scene keeps a representation that can be relit. RxGS preserves cross-view geometric consistency and produces illumination-dependent appearance under novel lighting.

\section{Statements and Declarations}

\subsection{Competing Interests}
The authors declare that they have no competing interests.

%% The sn-mathphys documentclass option already sets the bibliography style;
%% an explicit \bibliographystyle would emit a duplicate \bibstyle and break bibtex.
\bibliography{reference}

\end{document}